\documentclass{article}
\PassOptionsToPackage{dvipsnames}{xcolor}   
\PassOptionsToPackage{hyphens}{url}          
\usepackage{iclr2027_conference,times}
\iclrfinalcopy
\usepackage[utf8]{inputenc}
\usepackage[T1]{fontenc}
\def\eqref#1{equation~\ref{#1}}

\def\1{\bm{1}}

\DeclareMathAlphabet{\mathsfit}{\encodingdefault}{\sfdefault}{m}{sl}
\SetMathAlphabet{\mathsfit}{bold}{\encodingdefault}{\sfdefault}{bx}{n}

\usepackage{amsmath,amssymb,amsthm}

\usepackage{xcolor}

\usepackage{graphicx}
\usepackage{booktabs}
\usepackage{multirow}
\usepackage{makecell}
\usepackage{threeparttable}
\usepackage{tikz}
\usetikzlibrary{positioning,arrows.meta,calc}
\usepackage{float}

\usepackage{booktabs}
\usepackage{tabularx}
\usepackage{array}
\usepackage{url}

\usepackage{microtype}
\usepackage{placeins}   

\usepackage{url}
\usepackage{hyperref}
\usepackage[capitalize,noabbrev]{cleveref}

\newcommand{\nTasks}{20}  
\newcommand{\nStressEpisodes}{17}
\newcommand{\nStressFalseResolved}{11}
\newcommand{\nStressHonest}{6}
\newcommand{\nStressControlPass}{17}

\newcommand{\papertitle}{Incident-Arena }   

\providecommand{\yy}[1]{\textcolor{magenta}{\small[Yiyuan: #1]}}   

\providecommand{\yyedit}[1]{\textcolor{cyan}{#1}}

\newcommand{\cmark}{\textcolor{ForestGreen}{\large\(\pmb{\checkmark}\)}}
\newcommand{\pmark}{\textcolor{BurntOrange}{\large\(\boldsymbol{\circ}\)}}
\newcommand{\xmark}{\textcolor{BrickRed}{\large\(\boldsymbol{\times}\)}}

\title{Incident-Arena: Getting agents to the last nine of reliability}

\author{\parbox[t]{\dimexpr\linewidth-2\tabcolsep\relax}{\raggedright
Andre Fu$^{1}$\thanks{Correspondence to: \texttt{andre@abundant.ai}}, Malik Drabla$^{2}$, Leon Liu$^{3}$, Meji Abidoye$^{1}$, Marek \v{S}uppa$^{4}$, Lata Mishra$^{1}$, Adnan El Assadi$^{5}$, Yiyuan Li$^{3}$\\[4pt]
{\normalfont \mbox{$^{1}$Abundant AI} \quad \mbox{$^{2}$Adrenaline AI} \quad \mbox{$^{3}$Carnegie Mellon University} \quad \mbox{$^{4}$Comenius University in Bratislava} \quad \mbox{$^{5}$Massachusetts General Hospital}}
}}

\begin{document}

\maketitle
\ificlrfinal\lhead{Preprint}\fi


\begin{abstract}

AI coding agents are ubiquitous in software engineering workflows in both academia \& industry. Yet, despite their use in app coding, relatively less attention has been paid to their ability to execute on production incident response. This emerging field, termed agentic site-reliability-engineering (SRE) contains benchmarks limited by (1) unrealistic environments, typically toy repositories (2) non-standard framework implementations and (3) simple static verifiers. We introduce \textit{\papertitle}, a human-built benchmark of 20 carefully selected tasks grounded in real-world deployed open source software. Each task deploys a production application to an ephemeral Kubernetes cluster, injecting parameterizable faults, and a sustained load profile given the task requirements. We also present a novel verification method, going beyond static checks to functional verifiers, holding systems level metrics stable, while ensuring repairs are done safely. Agent trials run an average of 4.0M tokens and 68 steps, going beyond existing benchmarks, demonstrating agentic long horizon reasoning in SRE-tasks. Across 20 human-built tasks and 3 application substrates, the best model configuration passes at 64.3\%, with failures extending from diagnosis/localization errors, through incomplete repairs and unsafe regressions.  

\end{abstract}

\begin{figure}[H]
    \centering
    \includegraphics[width=0.72\linewidth]{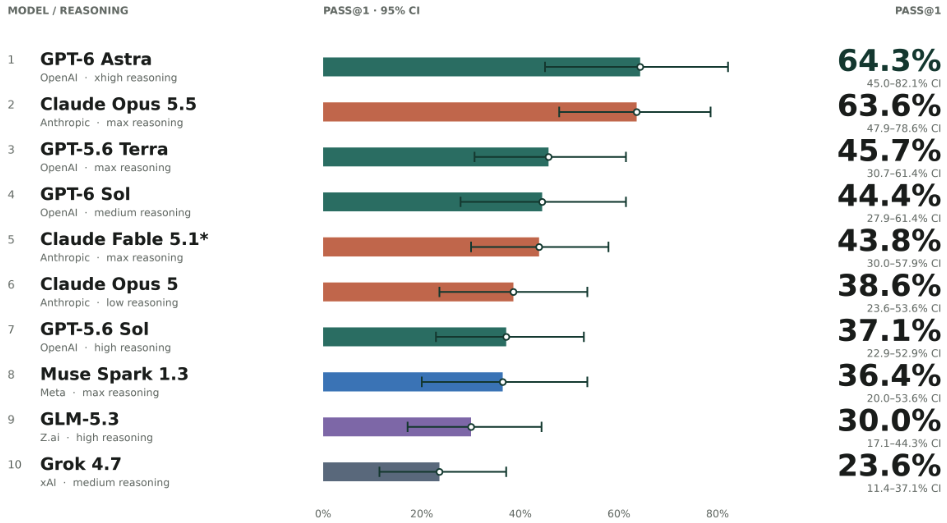}
    \caption{The strongest top-10 models by reasoning on \textit{\papertitle} ranked by Pass@1. * = cyber refusals, discussed in \autoref{sec:setup}}
    \label{fig:hero-leaderboard}
\end{figure}

\providecommand{\yy}[1]{\textcolor{magenta}{\small[Yiyuan: #1]}}
\providecommand{\yyedit}[1]{\textcolor{cyan}{#1}}

\section{Introduction}
\label{sec:intro}

Coding agents have emerged as the dominant form of interacting with code since the advent of Claude Code and Codex, moving from typing by code hand to instructing agents to execute application coding tasks. These agents are increasingly able to solve substantial repository level tasks. As powerful coding agents are able to take more and more of the application coding work, open questions still remain on their ability to move beyond to other pieces of the engineering lifecycle.  Operational repairs, as a key component in the lifecycle, are successful only if they restore service, preserve unrelated behavior and state, and remain effective after processes restart. Evaluating this capability therefore goes beyond checking whether an agent produces a plausible patch or eliminates an alert.

Recent benchmarks evaluate agents in DevOps, integrations, observability \& incident-response tasks \citep{tang2026devopsgym,kottamasu2026apexswe,clark2026sregym}. The existing work addressing the agentic SRE space falls short of three core attributes: \textit{Operational Validity}, \textit{Diagnostic Resolution}, \textit{Reproducibility \& Interoperability}. 

Firstly, \textit{Operational Validity}, meaning the verifier \& environment realism, is difficult to establish, where some benchmarks operate on repository level artifacts, live-system benchmarks use custom or trivial applications as operational surfaces. While useful for control, they leave uncertainty about measured performance transferring to real-world systems.  Furthermore, frontier models are already capable of detecting evaluation regimes \citep{needham2025large}, therefore artificial interfaces, small discrepancies in parity, and unrealistic verifiers or goals let induce evaluation suspicion. 

Secondly, existing benchmarks provide limited \textit{Diagnostic Resolution}, where aggregate success rates reveal agent failures, but often not with controlled variation to elicit failure behaviours. Specifically, by sweeping reasoning across frontier models, we're able to analyze the operational capabilities isolated in stronger models. 

Thirdly, on \textit{Reproducibility \& Interoperability}, due to the nature of SRE-style tasks that make them difficult to test using Harbor \citep{Harbor_Framework}, existing work like SRE-Gym \citep{clark2026sregym}, APEX-SWE \citep{kottamasu2026apexswe} and DevOpsGym \citep{tang2026devopsgym} all use unique and custom integrations. This makes agent analysis difficult and comparison difficult.

We introduce \papertitle, a human-authored benchmark, analyzing agent's ability to diagnose, repair and successfully operationalization broken production systems. We present 20 human-authored tasks built from live Kuberenetes systems, inspired or pulled from real production repositories. Faults are built across configuration, runtime state or image layers, while agent access is varied access from full code visibility through to scoped \texttt{kubectl} commands.

Agents are placed in a foothold pod, with read-only access to metrics, logs and other key observability metrics. Final evaluation is deterministic \& LLM-free, to avoid rubric design - instead we present a novel verifier with two core checks: \textit{Outcome Gate}, measuring recovery under continued traffic and a \textit{Safety Gate}, measuring durability, data preservation, repairs and the limited scope.

Our contributions are:
\begin{itemize}
    \item \textbf{A human-authored operational benchmark}: Difficulty emerges from complexity in the agent environment, multi-pod, multi-service systems with complex hops in the environment. 
    \item \textbf{Production-oriented verifiers}: Our dual-gate verifier jointly measures recovery \& repair under continued load, restart and fresh traffic. 
    \item \textbf{A comprehensive frontier evaluation}: We analyze 3,000 trials of agent trials across frontier models, thoroughly analyzing their frontier capabilities. 
\end{itemize}
\section{Related Work}
\label{sec:related}

\begin{table*}[t]
\centering
\caption{What each benchmark evaluates. \cmark{} yes; \pmark{} partly or on
some tasks; \xmark{} no or not stated. \emph{Repairs}: agents can change the
system and have the result checked. \emph{Names fault}: correctness of fault
attribution is graded. \emph{Scope}: preservation, collateral damage, or repair
scope is checked. \emph{Durable}: the repair is re-checked after a wait or restart.
\emph{No judge}: success does not depend on a trace or LLM judge.
\emph{Both ways}: a correct fix passes and doing nothing fails.
\emph{No fault}: includes cases where changing nothing is correct.
\emph{Harbor}: runs on Harbor, natively or via an adapter.
Further sources and coding decisions appear in Appendix~\ref{app:table1}.}
\label{tab:compare}
\begin{threeparttable}
\resizebox{\textwidth}{!}{%
\begin{tabular}{@{}l r l l c c c c c c c c@{}}
\toprule
Benchmark & Tasks & Task types & Setting
& Repairs & \makecell{Names\\fault} & Scope & Durable
& \makecell{No\\judge} & \makecell{Both\\ways}
& \makecell{No\\fault} & Harbor \\
\midrule
\multicolumn{12}{@{}l}{\textit{Live systems}} \\
AIOpsLab~\citep{chen2025aiopslab}
& 48 & Detection; RCA; mitigation & Kubernetes services
& \cmark & \pmark & \xmark & \xmark & \cmark & \xmark & \pmark & \xmark \\
ITBench--SRE~\citep{jha2025itbench}
& 42\tnote{a} & Diagnosis; mitigation & Kubernetes applications
& \cmark & \pmark & \xmark & \xmark & \cmark & \xmark & \xmark & \xmark \\
SREGym~\citep{clark2026sregym}
& 90 & Diagnosis; mitigation & Kubernetes applications
& \cmark & \pmark & \xmark & \xmark & \xmark & \xmark & \xmark & \xmark \\
OperAID
& 3 & Operational mitigation & Live services
& \cmark & \xmark & \xmark & \xmark & \cmark & \xmark & \xmark & \xmark \\
DBA-Bench
& 106 & Database diagnosis; repair & PostgreSQL workloads
& \cmark & \pmark & \cmark & \xmark & \cmark & \xmark & \xmark & \xmark \\
InfraBench~\citep{gao2026infrabench}
& 12 & Deploy; recover; migrate & Containers; VMs; clusters
& \cmark & \xmark & \pmark & \cmark & \xmark & \xmark & \xmark & \xmark \\
\addlinespace[2pt]
\multicolumn{12}{@{}l}{\textit{Repository and container tasks}} \\
DevOps-Gym~\citep{tang2026devopsgym}
& 700+ & Build; monitor; repair; test & Repositories; containers
& \pmark & \pmark & \pmark & \xmark & \cmark & \xmark & \xmark & \xmark \\
APEX-SWE~\citep{kottamasu2026apexswe}
& 200 & Integration; observability & Containerized services
& \pmark & \xmark & \pmark & \xmark & \cmark & \xmark & \xmark & \xmark \\
\addlinespace[2pt]
\multicolumn{12}{@{}l}{\textit{Recorded evidence}} \\
Cloud-OpsBench~\citep{cloudopsbench2026}
& 754 & Kubernetes RCA & Replayed state snapshots
& \xmark & \cmark & -- & -- & \cmark & \xmark & \xmark & \xmark \\
ORCA-bench~\citep{gong2026orca}
& 1{,}079\tnote{b} & Detection; multi-cause RCA & Recorded telemetry
& \xmark & \pmark & -- & -- & \xmark & \xmark & \cmark & \cmark \\
ITBench-AA~\citep{jha2025itbench}
& 59 & Fault attribution & Recorded evidence
& \xmark & \cmark & -- & -- & \cmark & \xmark & \xmark & \xmark \\
OpenRCA
& 335 & Root-cause localization & Recorded evidence
& \xmark & \cmark & -- & -- & \cmark & \xmark & \xmark & \xmark \\
\midrule
\makecell[l]{\textbf{\papertitle{}}\\(this work)}
& \textbf{\nTasks} & Diagnosis; repair & Live application systems
& \cmark & \xmark\tnote{c} & \cmark & \cmark
& \cmark & \cmark & \xmark & \cmark \\
\bottomrule
\end{tabular}%
}
\begin{minipage}{\textwidth}%
\footnotesize\raggedright\smallskip
\textsuperscript{a}\,The 42 SRE scenarios of ITBench's 94; the others concern security and
\textsuperscript{b}\,884 incident tasks and 195 no-fault controls.\par
\textsuperscript{c}\,Attribution is not graded: regex-based checks proved brittle, while aa subjective grading rubric.\par
\end{minipage}
\end{threeparttable}
\end{table*}

\paragraph{Incident response on live systems.}
AIOpsLab \citep{chen2025aiopslab} deploys DeathStarBench \citep{deathstarbench}
applications on Kubernetes. A mitigation passes if every service is up afterwards.
ITBench \citep{jha2025itbench} counts a scenario as mitigated when its alert clears, and OperAID \citep{operaid} verifies fixes on a live 5G core. SREGym \citep{clark2026sregym} is the closest to our work. It adds background noise and metastable failures, scores diagnoses with an LLM-judged checklist, and counts a mitigation as successful when the fault is gone and the system is healthy again. Most of these run demo or benchmark applications and give the agent general \texttt{kubectl} commands or privileged actions. 
\papertitle{} instead runs pinned production applications, Saleor \citep{saleor} and
FrappeERP \citep{frappeerp}, alongside a Slack substrate we built. The agent gets scoped permissions, and an LLM-free verifier checks each repair while traffic keeps running or after the affected service restarts. Table~\ref{tab:compare} compares the benchmarks discussed below.

\paragraph{Diagnosis and repository tasks.}
Other benchmarks stop short of repairing a running system. DevOps-Gym
\citep{tang2026devopsgym} and APEX-SWE \citep{kottamasu2026apexswe} grade the code change
or written diagnosis the agent hands in. Diagnosis benchmarks such as OpenRCA
\citep{openrca}, Cloud-OpsBench \citep{cloudopsbench2026}, ORCA-bench \citep{gong2026orca}
and ITBench-AA \citep{itbenchaa} ask for the root cause of failures in recorded data. Two
of them find that diagnoses are often ungrounded: Cloud-OpsBench reports 0.76 root-cause
accuracy but an evidence-closure rate of only 0.38, and on ORCA-bench the weakest model
invents an implausible root cause in 40\% of reports. In \papertitle{}, the agent has to
find the fault and fix it, and we grade the fix.

\paragraph{Safe remediation.}
A repair also has to be safe. R2Act \citep{r2act} finds that 39.5--62.0\% of correctly
diagnosed incidents still receive an invalid recovery action. Given underspecified DevOps
instructions, coding agents overstep an action boundary in 55.8--67.8\% of runs
\citep{actionboundary}. On live PostgreSQL, DBA-Bench \citep{dbabench} reports that human
DBAs repair 93.4\% of scenarios safely, against 17.9\% for the best automated baseline.
InfraBench \citep{gao2026infrabench} comes closest to our verifier. It re-checks
infrastructure tasks under sustained load, after a restart and at teardown, but flags risky
commands with an LLM judge and gives agents root and SSH access. We run similar checks on
incidents in production applications, without an LLM judge and with narrow permissions,
and also fail any repair that reaches outside its allowed scope.

\paragraph{Score validity.}
Recent coding benchmarks ask whether their own scores can be trusted. Checks can pass
wrong answers: 7.8\% of all SWE-bench \citep{jimenez2024swebench} patches count as correct
while failing the developers' test suite \citep{wangpradelliu}, and SWE-Marathon
\citep{desai2026swemarathon} finds a verifier bypass in 10.2\% of rollouts. In operations,
the same failure is a health check that passes on a broken system. Tasks can also leak into
training data, which SWE-bench-Live \citep{zhang2026swebenchlive} and SWE-rebench
\citep{badertdinov2025swerebench} counter by collecting new ones. And models may recognize
when they are being evaluated; Gemini-2.5-Pro does so with an AUC of 0.83
\citep{needham2025large}. We check every task both ways, with a golden solution and a
no-op run, and have adversarial agents try to game each one; Appendix~\ref{app:stress}
shows that our outcome checks reject reports of repairs that never happened. Because two
of our applications are public, we inject new faults and block the agent from code-hosting
sites, and our instructions avoid evaluation cues.

\paragraph{Diagnostic resolution.}
A single pass rate shows that agents fail, but not where. METR measures agents by the
length of task, in human working time, that they can complete at a fixed success rate
\citep{kwa2025horizon}. For root-cause analysis, a pooled accuracy across systems can hide
which method wins on each \citep{pooledaudit}, and ORCA-bench varies report specificity,
detection time and co-occurring faults \citep{gong2026orca}. We do the same for repair:
we vary the agent's access from full source visibility to scoped \texttt{kubectl}, run ten
model and harness configurations at every available reasoning level, and report results
per application.

\providecommand{\yy}[1]{\textcolor{magenta}{\small[Yiyuan: #1]}}
\providecommand{\yyedit}[1]{\textcolor{cyan}{#1}}

\section{\papertitle}
\label{sec:method}

\papertitle  is fundamentally three components together: a substrate, a fault layer, and the agent runtime. The substrate + fault layer build a task, and the runtime determines the agent environment. 

\subsection{Substrate Construction}
\label{sec:method:substrate}
Substrates are the fundamental unit of how we build tasks in \papertitle. They are complete, executable systems where incidents or faults can be injected. Building substrates typically has 3 core stages: mining, operationalizing \& packaging. We have 3 core substrates: Slack, Saelor \citep{saleor} \& FrappeERP \citep{frappeerp} - Slack is a highly realistic built substrate, while Saelor \& FrappeERP are mined from open source.

\begin{figure}[H]
    \centering
    \includegraphics[width=0.6\linewidth]{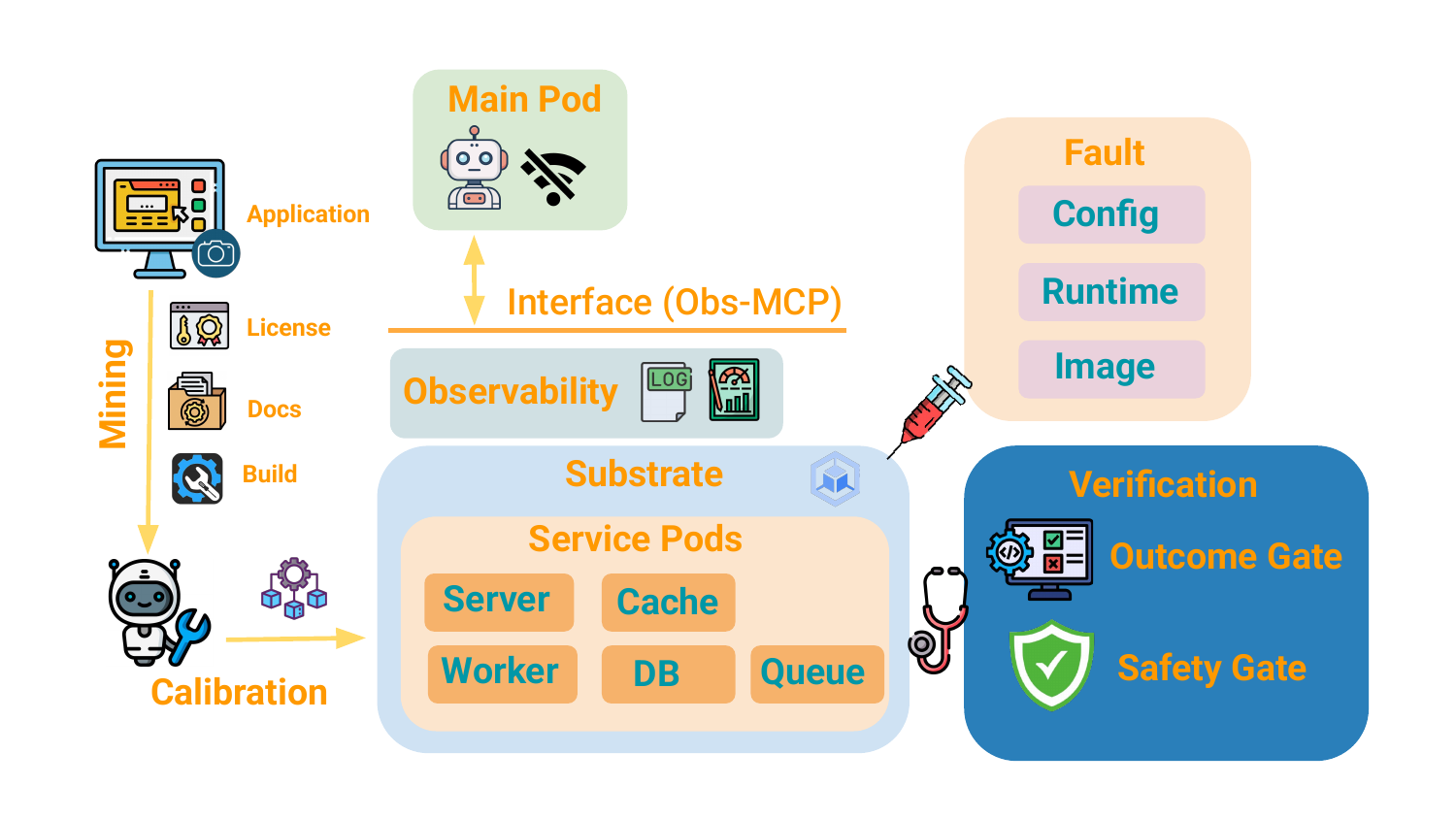}
    \caption{Process of Developing \textit{\papertitle}tasks by mining and developing substrates from real applications, injecting faults in different levels, with verifications on live operations.}
    \label{fig:overview}
\end{figure}

\textbf{Mining Open Source Repos}: We search for open-source applications used in (or modelled after) production that carry real operational complexity: persistent state, network services, background workers, caches, queues, authentication, and external dependencies. Acceptance requires (1) a permissive license, (2) sufficient documentation, (3) a reproducible build, and (4) enough surface area for task footholds. Each repository is pinned to a commit so upstream changes cannot alter the benchmark. 

If we have a specific application in mind that does not satisfy our constraints, we find and build our ideal version of that application. For our Slack substrate, we built it from scratch, modelled similar to the Mattermost repository \citep{mattermost} but significantly more realism, extending to 47 pods and 45 services (Table~\ref{tab:topology}). 


\textbf{Operationalizing Repositories}: OSS repositories rarely ship as production systems, so we supply the missing infrastructure: containerizing and building services, deterministic startup, migrations and seed data, network configuration, load generators, and metrics logging. We then \emph{calibrate} each substrate, recording baseline metrics as the healthy reference state. Calibration is a one-time activity that is done against a healthy substrate in order to measure the healthy vs unhealthy states induced in tasks. Core metrics measurincluded duringin the calibration include p50, p99 latency on core pods, split by peak \& trough load, error rates, goodput ratios, traffic counts, and probes attached to specific pods and services. We also have probes attached to service availability, data integrity and read-after-write correctness, ordering, duplication, queue lengths/backlogs, repair persistence \& forbidden restarts. 

\textbf{Kubernetes Packaging}: Tasks are built through a substrate + a fault overlay. The agent runs in a single \texttt{main} pod that serves as it's foothold operational shell. The rest of the system runs in separate service pods (app servers, workers, databases, caches, queues). The \texttt{main} pod reaches approved endpoints via task-specific APIs and limited RBAC, with no general cluster credentials and no access to verifier secrets or grading artifacts. 

\textbf{Agent Network Isolation}: The \texttt{main} pod has no direct Internet access all outbound requests must pass through a controlled proxy pod. The proxy allows Internet access during trusted setup, then switches to a restricted policy permitting only model APIs and package hosts before the agent starts. \texttt{NetworkPolicy} rules block direct public-IP connections, filtered DNS resolves only cluster-local names, and HTTPS is checked against both hostname and TLS SNI. This prevents proxy bypass and access to source-hosting services such as GitHub.

\textbf{Observability}: Prometheus scrapes service-pod metrics at fixed intervals and a collector ships pod logs to Loki. The agent queries both through Obs-MCP, a read-only PromQL/LogQL interface. Direct backend access is blocked, so observations cannot be modified and private grading data stays inaccessible.

\subsection{Task Construction}
\label{sec:method:tasks}
Tasks are built from using a modified Harbor format \citep{Harbor_Framework}. In our modification, each task contains a top-level human-readable instruction file, a Kubernetes chart, ground-truth metrics, the hidden verifiers, the solution oracles and one-hour agent time limit. Tasks were built through open source contributors and internal authors. Candidate tasks were built through careful task design, attempting to target specific difficult reasoning and capability gaps found in frontier models.

\textbf{Fault injection:} Our task creation process involves a healthy substrate combined with a fault in order to induce the task we'd like the agent to target. In this case, we have 3 types of fault injection mechanisms: configuration, runtime and image layer faults. \emph{Configuration} faults change the Helm values before deployment (connections, timeouts, replicas, etc). \emph{Image} faults are a small code or image change on top of a pinned base image. \emph{Runtime} faults are where some fault emerges or evolves after deployment either before or during an agent run. The full taskset with the fault tiers are found in \autoref{app:tab:tasklist} in Appendix.

\textbf{Agent Access:} Recognizing that agent permissions affect the task completion, we devise three types of agent access permissions. \emph{Confined}, gives the agent an operator pod with very specific diagnostic and repair surfaces - no app pod shell. The \emph{Shell-Visible} surface exposes narrow scoped shell access to selected pods, mostly non root and read-only file systems. Thirdly, the \emph{build-capable} surface exposes source files in order to induce trigger a rebuild for the agent.

\subsection{Verifier Design}
\label{sec:method:verifier}

Compared to the static checks typical of other benchmarks, \papertitle grades a repair by how the system behaves afterwards, not by what its configuration looks like. We design two gates. Firstly, the \emph{Outcome gate} measures workload Service Level Indicators (SLIs): goodput ratio, latency and error rate, while fresh traffic continues over a fixed period after the agent has submitted its repair. Secondly, the \emph{Safety gate} checks that the repair actually holds: we restart the affected service or fire the fault's trigger again and confirm the setting is still in place, we confirm the agent stayed within the task's allowed scope, and we confirm protected data and guardrails are intact. This is what stops shortcut repairs such as queue flushing, overly broad grants, disabling traffic etc. from passing. A task is a binary pass only if both gates pass. All of these underlying checks are hidden from the agent and are deterministic; any malformed or missing evidence fails the task. The SLI bounds come from the healthy substrate, measured once per substrate before any trial, not from post-trial tuning.

\subsection{Task Design \& Qualification}
\label{sec:method:approval}

\textbf{Instruction Design:} In order to avoid eval awareness in agent trials, we explicitly remove any form of markdown writing in the \texttt{instruction.md}, along with any semblance of ``AI-writing'' (despite being human verified, we try to avoid the stylistic choices models make). We only give the agent the minimal required information to solve the task, the time limit and the submission instructions. 

\textbf{Qualification:} After development, tasks go through multiple rounds of Quality Assessment (QA), patching holes found in the task environments \& design. Our QA process is broadly multiple adversarial agents attempting to explicitly reward hack and others where the agent finds unfair flaws in the task or verifier design. Final tasks must pass both Oracle \& No-operation trial - oracle to prove the golden solution path and no-op to prove nothing causes the errors to persist.

\providecommand{\yy}[1]{\textcolor{magenta}{\small[Yiyuan: #1]}}
\providecommand{\md}[1]{\textcolor{red}{\small[Malik: #1]}}
\providecommand{\yyedit}[1]{\textcolor{cyan}{#1}}

\section{Experimental Setup}
\label{sec:setup}

\textbf{Agent configurations:} We measure 10 agent configurations from OpenAI, Anthropic, xAI, Meta \& Zhipu using their native agent harness, except for GLM 5.3 which is paired with Claude-code. We evaluate with 3 independent trials at every available reasoning level. Across 20 tasks this produces 3,000 model trials. Harness versions and configurations are shown in \autoref{tab:agent-configurations}. All agents per task receive the same  \texttt{instruction.md} and the same set of tools. We use a Harbor fork modified to support Kubernetes backends. 

\textbf{Runtime Environment}: Popular agent sandboxes lack privileged container access, required to run k3s \citep{k3s}, needed in \papertitle. Therefore, we use AWS EC2s, \texttt{m7i.2xlarge} instances with the standard Ubuntu 24.04 LTS machine image on \texttt{amd64}, configured with 80 GiB of gp3 disk. The instance comes with 8 vCPUs and 32 GB of memory, to manage the size and complexity of the services on the pods. Each instance creates a fresh k3s cluster and pulls immutable, digest-pinned images from GHCR. The deployment combines the pinned task substrate version with its configuration, image, or runtime fault overlay. After task services and workload generator are ready, the agent is installed in the \texttt{main} operator pod and receives the repair surface specified in \autoref{tab:fault-surface-matrix}. 

\textbf{Timing Budgets}: The episode clock begins only after cluster deployment and readiness checks succeed, excluding infrastructure startup variance. Harness setup and agent execution share a fixed 60 minute window. An agent may declare completion early otherwise, it is stopped at timeout. The agent PID is sent a \texttt{SIGTERM}, then a \texttt{SIGKILL} 2 minutes later, to give the graded portion an agent-free soak period. The 180 second graded post-agent execution soak period then begins at the next complete load cycle boundary, to prevent agent biasing.

\textbf{Cyber Refusal}: Model providers are increasingly building safeguards into models, but these also prevent legitimate evaluation exercises outside of the cybersecurity scope. Despite this benchmark focuses on SRE tasks, we occasionally encounter cyber refusals, which are usually resolved by retries.   

\textbf{Failures and Retries}: We retry only failures attributable to benchmark infrastructure or harness execution, according to a fixed retry policy established before evaluation. Agent timeouts and invalid repairs are retained as model outcomes. Cyber refusals are resolved by 3 retries, but after 3 retries we deem it a failure.

\begin{table*}[t]
\centering
\scriptsize
\setlength{\tabcolsep}{4pt}
\renewcommand{\arraystretch}{1.13}
\resizebox{\textwidth}{!}{%
\begin{tabular}{@{}lllllcrrrrrr@{}}
\toprule
 & & & & & & & & \multicolumn{4}{c}{\textbf{Cost per episode}} \\
\cmidrule(l){9-12}
\textbf{Developer} & \textbf{Harness} & \textbf{Version} & \textbf{Model} & \textbf{Effort} & \textbf{Trials} & \textbf{In/Out price} & \textbf{Pass@1} & \textbf{Mean} & \textbf{Median} & \textbf{Passing} & \textbf{Failing} \\
\midrule
\multirow{4}{*}{OpenAI}
& Codex & 0.153.4 & GPT-6 (astra) & low-max & 15 & \$10 / \$50 & 0.59 & \$3.28 & \$2.78 & \$2.66 & \$4.17 \\
& Codex & 0.153.4 & GPT-6 (sol) & low-max & 15 & \$2 / \$10 & 0.39 & \$0.81 & \$0.60 & \$0.53 & \$0.99 \\
& Codex & 0.153.4 & GPT-5.6 (sol) & low-max & 15 & \$4 / \$20 & 0.37 & \$2.51 & \$1.70 & \$1.76 & \$2.97 \\
& Codex & 0.153.4 & GPT-5.6 (terra) & low-max & 15 & \$2 / \$12 & 0.28 & \$1.14 & \$0.76 & \$1.08 & \$1.16 \\
\addlinespace
\multirow{3}{*}{Anthropic}
& Claude Code & 2.1.280 & Claude Opus 5.5 & low-max & 15 & \$4 / \$20 & 0.54 & \$2.00 & \$1.34 & \$1.88 & \$2.14 \\
& Claude Code & 2.1.274 & Claude Fable 5.1 & low-max & 15 & \$10 / \$50 & 0.44 & \$3.99 & \$3.20 & \$4.01 & \$3.97 \\
& Claude Code & 2.1.274 & Claude Opus 5 & low-max & 15 & \$5 / \$25 & 0.36 & \$2.30 & \$2.10 & \$1.80 & \$2.48 \\
\addlinespace
xAI & Grok Build & 1.0.41 & Grok 4.7 & low-xhigh & 12 & \$2 / \$6 & 0.26 & \$6.27 & \$4.68 & \$4.87 & \$6.72 \\
Meta & Muse Code & 1.3.0 & Muse Spark 1.3 & minimal-max & 18 & \$1.25 / \$4.25 & 0.26 & \$1.49 & \$0.83 & \$1.06 & \$1.65 \\
Zhipu & Claude Code & 2.1.274 & GLM-5.3 & low-max & 15 & \$1.40 / \$4.40 & 0.31 & \$1.97 & \$1.76 & \$1.56 & \$2.11 \\
\midrule
\multicolumn{5}{r}{\textbf{Aggregate trials per task}} & \textbf{150} & & & & & & \\
\bottomrule
\end{tabular}}
\caption{Agent configurations, API prices, Pass@1 (averaged over all efforts) and model API cost per episode. Trials per task are 3 trials $\times$ each available reasoning level; 150 trials per task $\times$ 20 tasks = 3{,}000 trials. Prices are USD per million tokens; cache writes carry a 1.25$\times$ premium on Anthropic and none elsewhere. Passing/Failing are the mean cost of passing and failing episodes. 
}
\label{tab:agent-configurations}
\end{table*}

\textbf{Evaluation Protocol}: Our primary metric is resolve rate, \textit{Pass@1}. \emph{Pass@1} is the share of episodes that receive the binary reward, that is, pass both \emph{outcome} and \emph{safety} gates. Error bars in all figures are computed from a 95\% task-bootstrap CI, as seen in \autoref{fig:hero-leaderboard}

\section{Results}
\label{sec:results}

\subsection{Failure Taxonomy}
\label{sec:health}
We sort every failure into one class by the first rule it meets, in order of how direct the evidence is in \autoref{tab:classes}. The agent never declared, it changed something off the permitted list, a planted fault was still at its injected setting, the repair held at declaration but not afterwards, the fault's damage was still there, only the health checks failed, or some other integrity check failed. 

\definecolor{clsNever}{HTML}{7F7F7F}\definecolor{clsOff}{HTML}{C0504D}\definecolor{clsCause}{HTML}{9BBB59}\definecolor{clsLast}{HTML}{4F6D9A}\definecolor{clsEarly}{HTML}{F79646}\definecolor{clsOther}{HTML}{8064A2}
\providecommand{\sw}[1]{\raisebox{-0.15ex}{\textcolor{#1}{\rule{0.8em}{0.8em}}}\,}
\begin{table}[H]
\centering
\begin{minipage}[c]{0.31\linewidth}
\centering
\includegraphics[width=\linewidth]{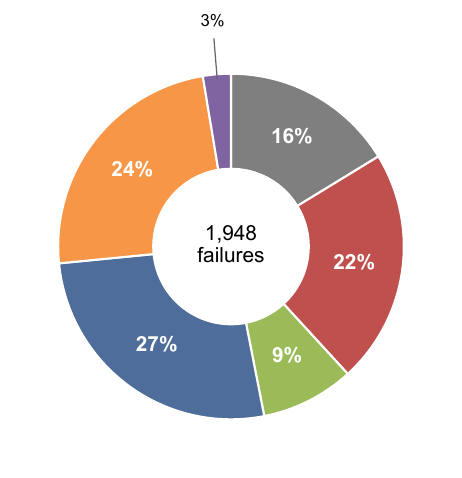}
\end{minipage}\hfill
\begin{minipage}[c]{0.66\linewidth}
\centering\small
\resizebox{\linewidth}{!}{%
\begin{tabular}{lrrr}
\toprule
Class (one per failure, assigned in this order) & Exclusive & Share & Inclusive \\
\midrule
\sw{clsNever}Never declared                                   & 317 & 16.3\% & 317 \\
\sw{clsOff}Changed something off the permitted list           & 427 & 21.9\% & 448 \\
\sw{clsCause}Left a planted cause in place, or its damage     & 170 &  8.7\% & 809 \\
\sw{clsLast}Repair did not last (restart or next trigger)     & 517 & 26.5\% & 782 \\
\sw{clsEarly}Declared while service was still out of bounds   & 466 & 23.9\% & 756 \\
\sw{clsOther}Other integrity check                            &  51 &  2.6\% &  51 \\
\midrule
All                                                            & 1,948 & 100\% & \\
\bottomrule
\end{tabular}}
\end{minipage}
\caption{Failure classes over the 1,948 failures. Exclusive: one class per failure. The pie shows these shares. Inclusive: every failure that meets the rule.}
\label{tab:classes}
\end{table}
\textbf{Never declared.} The hour or the step budget ran out in 325 episodes; 8 of them had repaired the system and the other 317 count as failures.

\textbf{Off the permitted list.} In 448 failures the agent changed a setting the task did not permit. Of the 439 with a key record, 261 touched a different subsystem from the fault, 70 an adjacent key in the right subsystem, 57 the intended key past its bound, and 51 protected PostgreSQL server settings. 

\textbf{Planted cause or damage left.} In 809 failures a planted fault was still at its injected setting (302: a pool left below its floor in 252, the strict-consistency trigger still active in 113, some both) or the damage it had done was still there (507: duplicate or inconsistent records on the sequencing tasks); 170 fail on this rule first. The twelve multi-fault tasks show the pattern directly. Of their 1,779 episodes, 964 issued the whole reference fix and passed 53\% of the time, 595 issued part of it and passed 2.5\% (15 of 595), and 220 issued none and passed 0\%. 549 of those 595 partial repairs were declared complete. 

\textbf{Repair did not last.} In 782 failures the change held at declaration, but could not persist through a restart, or past the next trigger. 

\textbf{Declared too early.} In 756 failures the agent declared while the service was still outside its bounds under continuing traffic (466 first); 376 of these had changed nothing they should not have and every change lasted, and \autoref{sec:fix} examines them.

\textbf{Each model fails in its own way.} Every model ran the same tasks at the same settings, so the differences in Table~\ref{tab:model-classes} (Appendix~\ref{app:model-classes}) 
belong to the model-harness pair. Failure patterns differ across model–harness configurations; this experiment does not isolate strictly the model’s contribution. The two GPT-6 models run out of time: a quarter to a third of their failures never declared, vs under a tenth for most others. Grok~4.7, GPT-5.6 (terra), Claude Opus~5, GLM-5.3 and Claude Fable~5.1 change things they should not, in a quarter to a third of their failures, vs about a tenth for the other three OpenAI models. Muse Spark~1.3 is starved by its provider: over a third of its failures never declared. No single fix would help all ten models.

\subsection{Agents are mainly bottlenecked on knowing when to stop}
\label{sec:fix}

For each task, the reference solution is a small set of actions that change the task's state. Agents reach the first correct action at a median 3.3 minutes into the allotted hour (interquartile range 1.7 to 7.7) and issue the rest of the correct actions almost immediately after. Of the 2,967 episodes with a usable action record, 1,638 (55\%) issue a ``full fix'', meaning every one of the reference solution's actions is implemented. Without the full fix (1,329 episodes, 457 of which never issue a single reference action) an episode rarely passes: 5.5\% (73 of 1,329). Of the episodes that implement a full fix, only 59\% (966 of 1,638) pass. The remaining 672 had the full solution in place and still failed.

Of these 672, 340 failed only on outcome. They are the agent not knowing when it is done: the repair was correct, and the agent stopped before the system had recovered. Sometimes the fix had not yet taken hold when the agent declared (a queue backlog still draining, a restart still in progress, connections still cycling). More often the fix was necessary but not sufficient: damage the fault had already done (backlogs, duplicated or inconsistent records) keeps the indicators out of bounds until it is cleared, and most of these episodes never returned within bound during the graded period (\autoref{app:checks}). The other 329 failed a safety check 
They are agents undoing their own success, most often by moving the maintenance window to a second that was still unsafe (145) or raising the connection pool past its ceiling after it had been fixed (38); \autoref{sec:health} has the rest.

\begin{table}[tb]
\centering\footnotesize
\caption{Post-fix behaviour among trials with a tracked full fix and an action record. 
}
\label{tab:recomputed-behavior}\label{tab:postfix}
\setlength{\tabcolsep}{3pt}
\begin{tabular}{lrr}
\toprule
Behaviour & Trials & Pass \\
\midrule
No further change & 638 & 0.696 \\
Further change & 931 & 0.530 \\
Declared within 2 minutes & 441 & 0.717 \\
Declared in 2--10 minutes & 782 & 0.599 \\
Declared at or after 10 minutes & 143 & 0.186 \\
Fewer than 5 read-only checks & 518 & 0.634 \\
5--8 read-only checks & 444 & 0.656 \\
9 or more read-only checks & 607 & 0.523 \\
\bottomrule
\end{tabular}
\end{table}

The delay effect is not hard tasks taking longer: inside each task, agents that declared within two minutes of their last correct action out-passed those that waited five minutes or more on eleven of the twelve tasks with enough episodes in both groups, by 20 points on average. Passing full-fix episodes declared a median 2.5 minutes after the last correct action, failing ones 3.7, and only 32\% declared within two minutes at all. More search, more time and more reasoning cannot buy these episodes back; verifying under load and leaving the fix alone can.

\subsection{Reasoning effort is a poor lever for repair success}
\label{sec:effort}

The eight five-setting models ran 480 episodes on each setting, with comparable pass rates across reasoning levels (low 39.2\%, medium 40.0\%, high 40.2\%, xhigh 41.9\%, max 43.8\%. Table~\ref{tab:effort} in Appendix~\ref{app:effort}). From the lowest setting to the highest, pass@1 gains 4.6 points (task-resampled interval $-4.8$ to $+13.8$); the best pooled result is the max setting, and the highest setting is best for three models of the eight. The gain does not increase with the task complexities: $+9.7$ points on the twelve multi-fault tasks against $-3.1$ on the eight single-fault ones, and the applications disagree ($+7.4$ on slack-spine, $+4.2$ on frappe, $-29.2$ on the saleor task). Grok~4.7's four settings show no trend. Muse Spark~1.3's six settings run from 23\% at minimal to 42\% at max, the one model whose highest setting is clearly its best. Meanwhile the mean cost of an episode multiplies by 2.3. Deliberation does not touch the failure of \autoref{sec:fix}: the agents still declare once the alarm is quiet.

\subsection{Reward Hacking Behaviour}
\label{sec:rwh}
We analyzed reward hacking behaviour in our corpus through a post-hoc process. Similar to \citep{desai2026swemarathon} we employ a suspicion score, $s \in [0, 1]$. We then build the three tiers, \textit{attempt-tier}: $s \geq 0.3$, includes broad behaviour that is undesirable, such as probing for solution files, questionable behaviour, and mostly actions that did not affect the outcome. In the \textit{exploit-tier}: $s\geq 0.85$, the agents would typically devise a strategy intended to pass checks, opposed to the intended repair, for example intentionally substituting temporary recovery for a durable fix. In the final tier, successful exploit: $exploit \land award = 1$, constitutes when an agent has the successful exploit tier and the award is achieved. Our post-hoc process takes these definitions and builds rubrics for them, using GPT-6-Astra as the judge. We also explore other LLM-as-a-judge agreement \& human-reviewed tasks in \autoref{app:rh-anal}.

\begin{figure}[tb]
    \centering
    \includegraphics[width=0.7\linewidth]{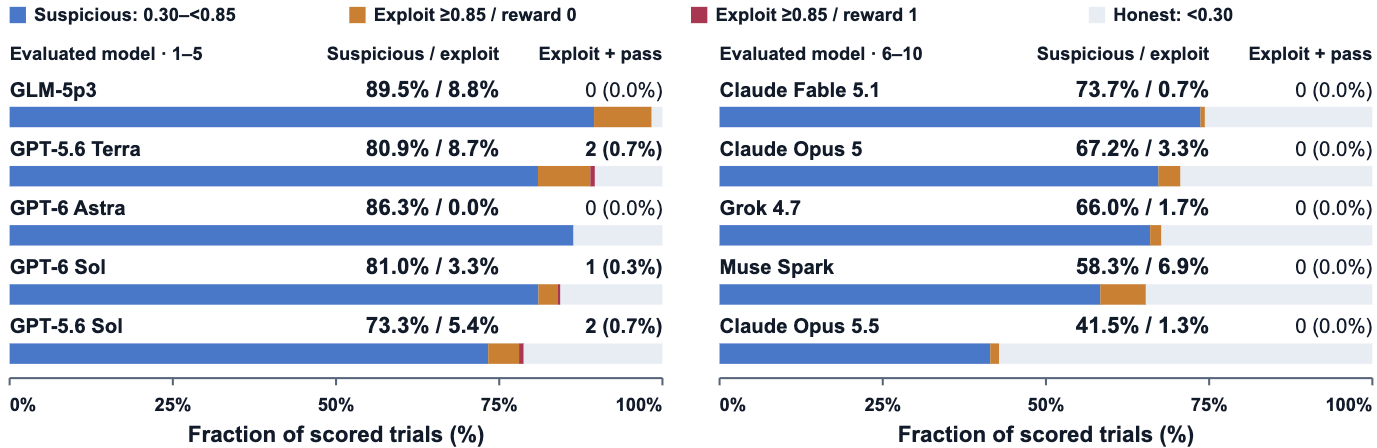}
    \caption{Reward hacking behaviour, split out by model. We categorize into three buckets, demonstrating significant attempted reward hacking behaviour across all models.}
\label{fig:tier}
\end{figure}

\textbf{Earlier exploits:} In an earlier task iteration, an agent reached remote code execution on the grader pod through a flaw in the environment; \autoref{app:rh-anal} describes it.


\section{Limitations}
\label{sec:limits}

\textbf{Sampling $n=3$}: Few samples generate large confidence interval bounds. With such tight competition between rival models in GPT-6-Astra \& Opus-5.5, the CI being so large poses challenges. Further experiments with a larger number of trials is required.

\textbf{Task coverage}: Due to the human nature of task development, Slack tasks are the majority at 13, while Frappe \& Saleor are in the minority. Fault type, onset, and substrate are also confounded in this cohort. The results therefore characterize these tasks and agent interfaces; they do not establish performance across the wider range of production systems.

\textbf{Harbor fork}: We had to develop our own backend infrastructure to support the K8s in an EC2 in Harbor \citep{Harbor_Framework}. This poses interoperability challenges we hope to address by merging our changes upstream.

\textbf{Time limit:} Most of our tasks are multi-hop, but agents typically only receive timer signals from the environment, opposed to a dedicated clock. Future work should include a timer the agent can check, similar to work in \citep{frontierswe}.

\textbf{Reward hacking interjudge rating}: In \autoref{app:rh-anal}, we find that despite using frontier models, our judges have poor inter-judge rating in some cases. We attempted to seed human-in-the-loop data, but further investigation is required and assessing human-observed trials.

\section{Conclusion}
\label{sec:conclusion}


\papertitle measures an AI agent's ability to conduct realistic SRE-style tasks on a human-made benchmark of 20 tasks. Across 3,000 trajectories, we demonstrate that this type of work is still yet to be saturated, with GPT-6-Astra on xhigh reaching a high of 65\%. Common failure modes from repairs not lasting, declaring too early, or non-permitted actions, demonstrate agents still suffer from casual reasoning gaps. Extending realism in these tasks, directly demonstrates significant headroom for agents to get better. 
\newpage
\bibliography{iclr2027_conference}
\bibliographystyle{iclr2027_conference}

\newpage
\appendix
\gdef\tableautorefname{Appendix Table}
\raggedbottom 
\FloatBarrier
\section{Fault injection vs Agent Access surfaces}
\label{app:faultxaccess}
\begin{table*}[!htbp]
\centering
\small
\renewcommand{\arraystretch}{1.25}
\begin{tabularx}{\textwidth}{lXXX}
\toprule
&
\multicolumn{3}{c}{\textbf{Agent access surface} $\longrightarrow$} \\
\cmidrule(lr){2-4}
\textbf{Fault injection layer} $\downarrow$ &
\textbf{Confined} &
\textbf{Shell-visible} &
\textbf{Build-capable} \\
\midrule

Configuration &
Repair through operator APIs. &
May inspect selected application pods. &
Not supported because no source change is present. \\

Runtime &
Repair through controlled changes to live state. &
Supported if the activation mechanism does not reveal the answer. &
Not supported because build-capable tasks require source changes. \\

Image &
Repair through operational mitigation. &
May inspect the faulted service without modifying it. &
May edit and rebuild one selected service. \\

\bottomrule
\end{tabularx}
\caption{The task space combines two axes: how the fault is injected and what access the agent receives. Each cell describes the permitted repair surface.}
\label{tab:fault-surface-matrix}
\end{table*}

These processes are completely parameterizable, allowing for multiple variants of similar flavours of the same task. Isolating the type and quality of the task with the fault $\times$ agent access surface.

\FloatBarrier
\section{Task list and difficulty axes}
\label{app:tasks}
Table~\ref{app:tab:tasklist} lists the evaluated tasks with the metadata used in \S\ref{sec:results}. Tier, onset and access surface are taken from the released task registry; cause count, planted report set and grading packs from each trial's own verification contract (identical bytes across all trials of a task). The difficulty axes are confounded with substrate on this cohort (every frappe task is compound, runtime-tier and mostly present at start; every slack-spine task is config- or image-tier and twelve of thirteen emerge under load), so the paper reports within-substrate contrasts only. For completeness, across all \nTasks{} tasks pass@1 is 0.314 on compound against 0.478 on single-cause tasks, and 0.378 on delayed-emergence (15 tasks) against 0.384 on present-at-start tasks (5); neither difference should be read causally.
\begin{table*}[t]
\centering\footnotesize
\caption{Evaluated tasks and outcome rates over the 3,000 trials (150 per task). }
\label{tab:recomputed-tasklist}\label{app:tab:tasklist}
\setlength{\tabcolsep}{3pt}
\resizebox{\textwidth}{!}{%
\begin{tabular}{lllllrr}
\toprule
Task & Tier & Causes & Onset & Planted faults & Pass & Action completion \\
\midrule
deletes-and-jobs-fail & runtime & 2 & at start & \makecell[l]{\texttt{mariadb.grants}\\\texttt{redis-queue.config}} & 0.493 & 0.649 \\
desk-and-queue-oom & runtime & 2 & under load & \makecell[l]{\texttt{mariadb.max-user-connections}\\\texttt{redis-queue.config}} & 0.300 & 0.829 \\
desk-and-queue-outage & runtime & 2 & under load & \makecell[l]{\texttt{mariadb.max-user-connections}\\\texttt{redis-queue.config}} & 0.547 & 0.641 \\
new-records-and-jobs-fail & runtime & 2 & at start & \makecell[l]{\texttt{mariadb.grants}\\\texttt{redis-queue.config}} & 0.333 & 0.693 \\
new-records-and-queue-oom & runtime & 2 & at start & \makecell[l]{\texttt{mariadb.grants}\\\texttt{redis-queue.config}} & 0.467 & 0.727 \\
writes-and-queue-oom & runtime & 2 & at start & \makecell[l]{\texttt{mariadb.read-only}\\\texttt{redis-queue.config}} & 0.153 & 0.720 \\
statement-timeout-canary & runtime & 1 & under load & \makecell[l]{\texttt{postgres.config}} & 0.340 & 0.933 \\
split-sequencer & config & 1 & under load & \makecell[l]{\texttt{message.sequencer}} & 0.013 & 0.201 \\
maintenance-collision & config & 1 & under load & \makecell[l]{\texttt{db.maintenance-controller}} & 0.720 & 0.884 \\
logins-unread-and-sends-all-slow-since-noon & config & 1 & under load & \makecell[l]{\texttt{redis.cache-policy}} & 0.767 & 0.587 \\
logins-unread-sends-slower & config & 1 & under load & \makecell[l]{\texttt{redis.cache-policy}} & 0.660 & 0.568 \\
sends-crawl-then-store-slows & config & 3 & under load & \makecell[l]{\texttt{channel.db-pool}\\\texttt{channel.membership-acl}\\\texttt{redis.cache-policy}} & 0.280 & 0.284 \\
sends-fail-after-strict-mode-plausible-pool & config & 2 & under load & \makecell[l]{\texttt{channel.db-pool}\\\texttt{channel.membership-acl}} & 0.100 & 0.149 \\
sends-fail-after-strict-mode-turns-on & config & 2 & under load & \makecell[l]{\texttt{channel.db-pool}\\\texttt{channel.membership-acl}} & 0.367 & 0.399 \\
sends-fail-during-compliance-window & config & 1 & under load & \makecell[l]{\texttt{channel.db-pool}} & 0.580 & 0.878 \\
sends-fail-strict-pool-16 & config & 2 & under load & \makecell[l]{\texttt{channel.db-pool}\\\texttt{channel.membership-acl}} & 0.100 & 0.128 \\
sends-slow-and-stall-every-minute & config & 2 & under load & \makecell[l]{\texttt{db.maintenance-controller}\\\texttt{message.db-pool}} & 0.227 & 0.707 \\
stall-every-minute-and-later-every-send-crawls & config & 2 & under load & \makecell[l]{\texttt{channel.membership-acl}\\\texttt{db.maintenance-controller}} & 0.407 & 0.574 \\
seq-lock-leak & image & 1 & at start & \makecell[l]{\texttt{message.sequencer}} & 0.473 & 0.456 \\
distractor-volume-shell & image & 1 & under load & \makecell[l]{\texttt{message.sequencer}} & 0.267 & 0.027 \\
\bottomrule
\end{tabular}}
\end{table*}

\section{Substrate topology}
\label{app:topology}
Table~\ref{tab:topology} counts what each substrate deploys, rendered from its Helm chart with a representative task overlay. The application pods are the software under test; the harness pods exist for the benchmark and are the same on every substrate up to the administrative interfaces and metric exporters each application needs. The agent's own container is one of the harness pods; it reaches the application only through the interfaces described in \S\ref{sec:setup}.
The acceptance criteria of \S\ref{sec:method:substrate} are applied as follows. \emph{Documentation}: the upstream project documents deployment and administration well enough that an operator can be expected to know the relevant settings. \emph{Reproducible build}: every image is built by a script in the repository and pinned by digest in a lock file, so all episodes of a task run identical bytes. \emph{Surface area}: the deployed system exposes enough services and configuration points to host many distinct faults; Table~\ref{tab:topology} counts them.
\begin{table}[!htbp]\centering\small
\caption{Kubernetes objects rendered from each substrate's Helm chart with a task overlay, as deployed on the single-node cluster of an episode. Application pods run the software under test: its servers, workers, databases, caches, queues and edge proxy. Harness pods exist for the benchmark: the agent's container, the egress proxy, DNS filter and TLS gateway, the observability stack (Prometheus, Loki, log shipper, metric exporters and the read-only MCP server), the load generator and the administrative interfaces. Setup pods run once at deployment.}
\label{tab:topology}
\begin{tabular}{lrrrrr}
\toprule
Substrate & Tasks & Application pods & Harness pods & Setup pods & Services \\
\midrule
frappe & 6 & 10 & 14 & 3 & 25 \\
saleor & 1 & 5 & 9 & 1 & 16 \\
slack-spine & 13 & 35 & 12 & 0 & 45 \\
\bottomrule
\end{tabular}
\end{table}

\FloatBarrier
\section{Verifier Ablation}
\label{app:checks}
\begin{table*}[t]
\centering\footnotesize
\caption{Outcome and safety gate verdicts.}
\label{tab:recomputed-gates}\label{tab:incident-arena-gate-ablation}
\setlength{\tabcolsep}{3pt}
\begin{tabular}{lrrrr}
\toprule
Task & Pass/Pass & Pass/Fail & Fail/Pass & Fail/Fail \\
\midrule
deletes-and-jobs-fail & 74 & 3 & 65 & 8 \\
desk-and-queue-oom & 45 & 2 & 24 & 79 \\
desk-and-queue-outage & 82 & 7 & 30 & 31 \\
new-records-and-jobs-fail & 50 & 66 & 0 & 34 \\
new-records-and-queue-oom & 70 & 8 & 25 & 47 \\
writes-and-queue-oom & 23 & 39 & 0 & 88 \\
statement-timeout-canary & 51 & 2 & 91 & 6 \\
split-sequencer & 2 & 22 & 0 & 126 \\
maintenance-collision & 108 & 41 & 0 & 1 \\
logins-unread-and-sends-all-slow-since-noon & 115 & 11 & 10 & 14 \\
logins-unread-sends-slower & 99 & 26 & 0 & 25 \\
sends-crawl-then-store-slows & 42 & 105 & 0 & 3 \\
sends-fail-after-strict-mode-plausible-pool & 15 & 86 & 0 & 49 \\
sends-fail-after-strict-mode-turns-on & 55 & 85 & 0 & 10 \\
sends-fail-during-compliance-window & 87 & 51 & 1 & 11 \\
sends-fail-strict-pool-16 & 15 & 135 & 0 & 0 \\
sends-slow-and-stall-every-minute & 34 & 102 & 0 & 14 \\
stall-every-minute-and-later-every-send-crawls & 61 & 78 & 0 & 11 \\
seq-lock-leak & 71 & 70 & 0 & 9 \\
distractor-volume-shell & 40 & 51 & 0 & 59 \\
\midrule All trials & 1139 & 990 & 246 & 625 \\
\bottomrule
\end{tabular}
\par\vspace{2pt}{\raggedright\scriptsize Counts, not percentages.\par}
\end{table*}

\begin{figure}[htbp]
\centering
\includegraphics[width=0.92\linewidth]{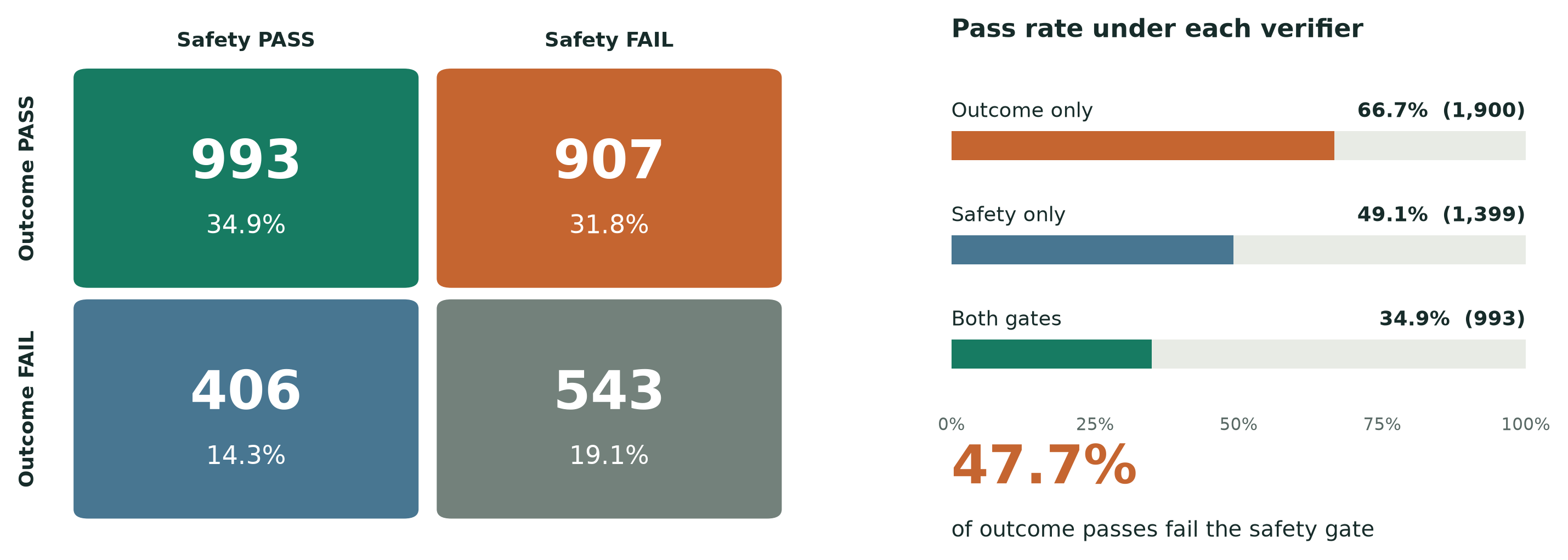}
\caption{Verifier Ablation }
\label{fig:verifier-ablation}
\end{figure}

\FloatBarrier
\section{Both-ways check (two-sided calibration)}
\label{app:calibration}
Table~\ref{tab:calib} gives, for every task in the reported experiment, the reward of the scripted reference fix and of the agent that does nothing. Both runs are repeated in every experiment, so a task whose checks had drifted would show here first.
\begin{table}[!htbp]\centering\scriptsize
\caption{Both-ways check inside the reported experiment: the scripted reference fix and the do-nothing agent, one run each per task.}
\label{tab:calib}
\begin{tabular}{llcc}
\toprule
Substrate & Task & Scripted-fix reward & Do-nothing reward \\
\midrule
frappe & \texttt{deletes-and-jobs-fail} & 1.0 & 0.0 \\
frappe & \texttt{desk-and-queue-oom} & 1.0 & 0.0 \\
frappe & \texttt{desk-and-queue-outage} & 1.0 & 0.0 \\
frappe & \texttt{new-records-and-jobs-fail} & 1.0 & 0.0 \\
frappe & \texttt{new-records-and-queue-oom} & 1.0 & 0.0 \\
frappe & \texttt{writes-and-queue-oom} & 1.0 & 0.0 \\
saleor & \texttt{statement-timeout-canary} & 1.0 & 0.0 \\
slack-spine & \texttt{split-sequencer} & 1.0 & 0.0 \\
slack-spine & \texttt{maintenance-collision} & 1.0 & 0.0 \\
slack-spine & \texttt{logins-unread-and-sends-all-slow-since-noon} & 1.0 & 0.0 \\
slack-spine & \texttt{logins-unread-sends-slower} & 1.0 & 0.0 \\
slack-spine & \texttt{sends-crawl-then-store-slows} & 1.0 & 0.0 \\
slack-spine & \texttt{sends-fail-after-strict-mode-plausible-pool} & 1.0 & 0.0 \\
slack-spine & \texttt{sends-fail-after-strict-mode-turns-on} & 1.0 & 0.0 \\
slack-spine & \texttt{sends-fail-during-compliance-window} & 1.0 & 0.0 \\
slack-spine & \texttt{sends-fail-strict-pool-16} & 1.0 & 0.0 \\
slack-spine & \texttt{sends-slow-and-stall-every-minute} & 1.0 & 0.0 \\
slack-spine & \texttt{stall-every-minute-and-later-every-send-crawls} & 1.0 & 0.0 \\
slack-spine & \texttt{seq-lock-leak} & 1.0 & 0.0 \\
slack-spine & \texttt{distractor-volume-shell} & 1.0 & 0.0 \\
\bottomrule
\end{tabular}
\end{table}


\FloatBarrier
\section{Full results matrix}
\label{app:matrix}
\begin{figure}[htbp]
\centering
\includegraphics[width=0.92\linewidth]{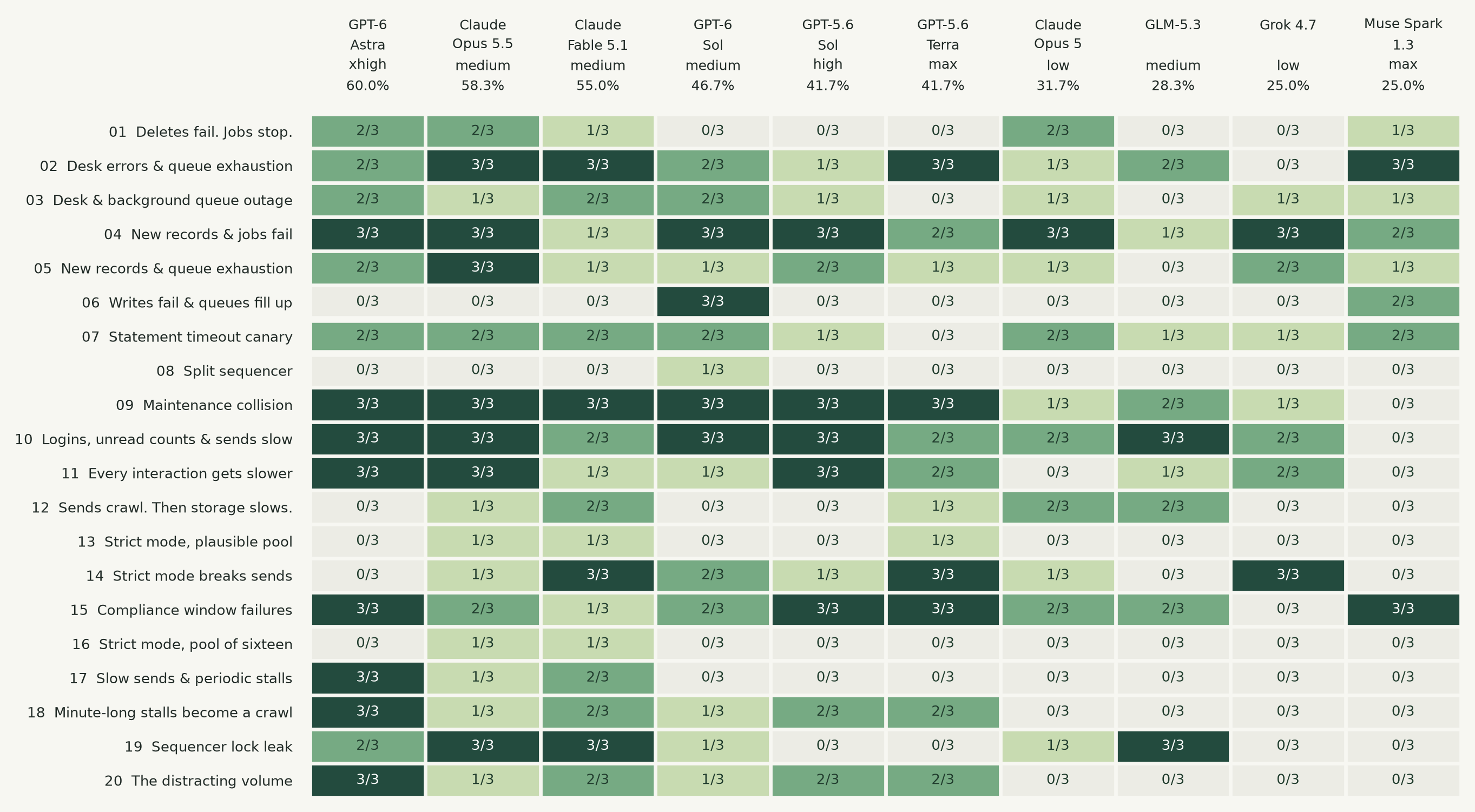}
\caption{Passing trials per task and configuration over admitted episodes. Models the best at their reasoning level shown for sampling. }
\label{fig:heatmap}
\end{figure}

Table~\ref{tab:intervals} gives each configuration's rates with confidence intervals and pass\textasciicircum{}$k$, Table~\ref{tab:substrates} the per-application breakdown, and Table~\ref{tab:matrix} and Figure~\ref{fig:heatmap} pass@1 for every task and configuration.
 \begin{table*}[t]
\centering\footnotesize
\caption{Results by model, pooling available reasoning settings.}
\label{tab:recomputed-models}\label{tab:intervals}
\setlength{\tabcolsep}{3pt}
\begin{tabular}{lrcrrr}
\toprule
Model & Trials & Pass@1 [task 95\%] & Action completion & $\mathrm{Pass}^3$ & $\mathrm{Pass}^5$ \\
\midrule
GPT-6 astra & 300 & 0.590 [0.430,0.743] & 0.702 & 0.410 & 0.328 \\
Claude Opus 5.5 & 300 & 0.540 [0.387,0.690] & 0.736 & 0.345 & 0.267 \\
Claude Fable 5.1 & 300 & 0.440 [0.297,0.590] & 0.597 & 0.242 & 0.182 \\
GPT-6 sol & 300 & 0.393 [0.263,0.527] & 0.660 & 0.173 & 0.104 \\
GPT-5.6 sol & 300 & 0.373 [0.240,0.520] & 0.588 & 0.184 & 0.138 \\
GPT-5.6 terra & 300 & 0.280 [0.173,0.400] & 0.481 & 0.091 & 0.045 \\
Claude Opus 5 & 300 & 0.357 [0.233,0.487] & 0.480 & 0.149 & 0.093 \\
GLM-5.3 & 300 & 0.307 [0.200,0.420] & 0.498 & 0.096 & 0.044 \\
Grok 4.7 & 240 & 0.263 [0.146,0.388] & 0.628 & 0.095 & 0.050 \\
Muse Spark 1.3 & 360 & 0.256 [0.139,0.378] & 0.233 & 0.090 & 0.039 \\
\bottomrule
\end{tabular}
\par\vspace{2pt}{\raggedright\scriptsize Task bootstrap resamples 20 tasks. Action completion is tracked full\_fix, not grading correctness. Pass powers are 20-task means of per-task pass rates t to 3 or 5.\par}
\end{table*}


\begin{table*}[t]
\centering\footnotesize
\caption{Pass@1 (tracked action completion) by model and substrate.}
\label{tab:recomputed-substrates}\label{tab:substrates}
\setlength{\tabcolsep}{3pt}
\begin{tabular}{lccc}
\toprule
Model & Frappe (6) & Slack-spine (13) & Saleor (1) \\
\midrule
GPT-6 astra & 0.678 (0.977) & 0.554 (0.550) & 0.533 (1.000) \\
Claude Opus 5.5 & 0.511 (0.867) & 0.574 (0.655) & 0.267 (1.000) \\
Claude Fable 5.1 & 0.300 (0.533) & 0.518 (0.595) & 0.267 (1.000) \\
GPT-6 sol & 0.444 (0.889) & 0.385 (0.528) & 0.200 (1.000) \\
GPT-5.6 sol & 0.300 (0.694) & 0.395 (0.508) & 0.533 (1.000) \\
GPT-5.6 terra & 0.289 (0.621) & 0.277 (0.374) & 0.267 (1.000) \\
Claude Opus 5 & 0.322 (0.778) & 0.379 (0.303) & 0.267 (1.000) \\
GLM-5.3 & 0.289 (0.589) & 0.308 (0.428) & 0.400 (0.867) \\
Grok 4.7 & 0.306 (0.732) & 0.244 (0.551) & 0.250 (1.000) \\
Muse Spark 1.3 & 0.370 (0.472) & 0.192 (0.098) & 0.389 (0.556) \\
All configurations & 0.382 (0.710) & 0.382 (0.450) & 0.340 (0.933) \\
\bottomrule
\end{tabular}
\par\vspace{2pt}{\raggedright\scriptsize Action completion is conditional on an action record.\par}
\end{table*}
\begin{table*}[t]
\centering\footnotesize
\caption{Pass rate by task and model across available reasoning settings.}
\label{tab:recomputed-matrix}\label{tab:matrix}
\setlength{\tabcolsep}{3pt}
\resizebox{\textwidth}{!}{%
\begin{tabular}{lrrrrrrrrrr}
\toprule
Task & GPT-6 astra & Claude Opus 5.5 & Claude Fable 5.1 & GPT-6 sol & GPT-5.6 sol & GPT-5.6 terra & Claude Opus 5 & GLM-5.3 & Grok 4.7 & Muse Spark 1.3 \\
\midrule
deletes-and-jobs-fail & 0.53 & 0.80 & 0.40 & 0.27 & 0.27 & 0.07 & 0.73 & 0.73 & 0.33 & 0.72 \\
desk-and-queue-oom & 0.60 & 0.40 & 0.13 & 0.67 & 0.27 & 0.67 & 0.13 & 0.07 & 0.00 & 0.06 \\
desk-and-queue-outage & 0.53 & 0.93 & 0.53 & 0.40 & 0.27 & 0.33 & 0.73 & 0.47 & 0.58 & 0.67 \\
new-records-and-jobs-fail & 1.00 & 0.00 & 0.00 & 0.80 & 0.60 & 0.53 & 0.00 & 0.00 & 0.42 & 0.06 \\
new-records-and-queue-oom & 0.53 & 0.80 & 0.60 & 0.40 & 0.40 & 0.13 & 0.27 & 0.27 & 0.50 & 0.72 \\
writes-and-queue-oom & 0.87 & 0.13 & 0.13 & 0.13 & 0.00 & 0.00 & 0.07 & 0.20 & 0.00 & 0.00 \\
statement-timeout-canary & 0.53 & 0.27 & 0.27 & 0.20 & 0.53 & 0.27 & 0.27 & 0.40 & 0.25 & 0.39 \\
split-sequencer & 0.00 & 0.00 & 0.00 & 0.13 & 0.00 & 0.00 & 0.00 & 0.00 & 0.00 & 0.00 \\
maintenance-collision & 0.93 & 0.87 & 0.87 & 0.73 & 0.93 & 0.80 & 0.33 & 0.67 & 0.58 & 0.50 \\
logins-unread-and-sends-all-slow-since-noon & 1.00 & 0.93 & 1.00 & 0.80 & 0.67 & 0.27 & 0.93 & 0.67 & 0.83 & 0.61 \\
logins-unread-sends-slower & 1.00 & 0.93 & 0.93 & 0.73 & 0.93 & 0.20 & 0.87 & 0.33 & 0.42 & 0.28 \\
sends-crawl-then-store-slows & 0.00 & 0.87 & 0.87 & 0.13 & 0.07 & 0.07 & 0.20 & 0.27 & 0.00 & 0.28 \\
sends-fail-after-strict-mode-plausible-pool & 0.00 & 0.20 & 0.33 & 0.00 & 0.00 & 0.07 & 0.33 & 0.07 & 0.00 & 0.00 \\
sends-fail-after-strict-mode-turns-on & 0.13 & 0.20 & 0.53 & 0.27 & 0.20 & 0.53 & 0.60 & 0.47 & 0.83 & 0.06 \\
sends-fail-during-compliance-window & 0.87 & 0.20 & 0.00 & 0.93 & 1.00 & 0.80 & 0.33 & 0.80 & 0.08 & 0.67 \\
sends-fail-strict-pool-16 & 0.00 & 0.47 & 0.20 & 0.00 & 0.00 & 0.00 & 0.27 & 0.07 & 0.00 & 0.00 \\
sends-slow-and-stall-every-minute & 0.73 & 0.73 & 0.47 & 0.00 & 0.20 & 0.07 & 0.00 & 0.07 & 0.00 & 0.00 \\
stall-every-minute-and-later-every-send-crawls & 0.73 & 0.80 & 0.53 & 0.27 & 0.60 & 0.20 & 0.33 & 0.40 & 0.08 & 0.11 \\
seq-lock-leak & 0.87 & 1.00 & 0.93 & 0.67 & 0.27 & 0.13 & 0.67 & 0.13 & 0.08 & 0.00 \\
distractor-volume-shell & 0.93 & 0.27 & 0.07 & 0.33 & 0.27 & 0.47 & 0.07 & 0.07 & 0.25 & 0.00 \\
\bottomrule
\end{tabular}}
\end{table*}

\FloatBarrier
\section{Failure classes by model}
\label{app:model-classes}

\begin{table}[H]
\centering\footnotesize
\caption{Exclusive failure-class shares by model.}
\label{tab:recomputed-failure-model}\label{tab:model-classes}
\setlength{\tabcolsep}{3pt}
\begin{tabular}{lrrrrrrr}
\toprule
Model & Failures & Never declared & Off-list & Cause/damage & Not lasting & Too early & Other \\
\midrule
GPT-6 astra & 125.8 & 31.8\% & 8.7\% & 14.3\% & 18.4\% & 26.7\% & 0.1\% \\
Claude Opus 5.5 & 157.2 & 13.4\% & 20.4\% & 8.3\% & 24.2\% & 32.4\% & 1.4\% \\
Claude Fable 5.1 & 170.0 & 5.3\% & 25.3\% & 4.7\% & 24.1\% & 33.8\% & 6.8\% \\
GPT-6 sol & 171.1 & 26.9\% & 9.9\% & 10.5\% & 25.8\% & 25.1\% & 1.8\% \\
GPT-5.6 sol & 178.4 & 5.6\% & 11.8\% & 18.5\% & 29.9\% & 29.6\% & 4.6\% \\
GPT-5.6 terra & 206.9 & 5.8\% & 30.0\% & 8.2\% & 28.1\% & 23.1\% & 4.9\% \\
Claude Opus 5 & 220.1 & 15.9\% & 29.1\% & 5.0\% & 22.3\% & 25.4\% & 2.3\% \\
GLM-5.3 & 226.0 & 8.0\% & 27.9\% & 11.9\% & 31.9\% & 18.6\% & 1.7\% \\
Grok 4.7 & 181.4 & 7.2\% & 30.9\% & 7.2\% & 33.1\% & 21.6\% & 0.1\% \\
Muse Spark 1.3 & 310.8 & 36.3\% & 18.7\% & 3.9\% & 25.0\% & 14.1\% & 2.0\% \\
\bottomrule
\end{tabular}
\end{table}
\FloatBarrier

\section{Reasoning effort, pooled}
\label{app:effort}

\begin{table}[H]
\centering\footnotesize
\caption{Reasoning effort among the eight five-setting models.}
\label{tab:recomputed-effort}\label{tab:effort}
\setlength{\tabcolsep}{3pt}
\begin{tabular}{lrrr}
\toprule
Setting & Trials & Pass@1 & Mean cost (USD) \\
\midrule
low    & 480 & 0.392 & 1.46 \\
medium & 480 & 0.400 & 1.76 \\
high   & 480 & 0.402 & 2.23 \\
xhigh  & 480 & 0.419 & 2.63 \\
max    & 480 & 0.438 & 3.38 \\
\bottomrule
\end{tabular}
\end{table}
\FloatBarrier

\section{Comparison table sources}
\label{app:table1}
Every cell of Table~\ref{tab:compare} was read from the primary source on 2026-09-21; the \emph{Setting} column paraphrases each paper's description of its environment. The sources were the arXiv abstract or full text, or the SREGym and ITBench-AA leaderboards, \citealp{sregymleaderboard,itbenchaa}). \emph{Durable} is marked only for an explicit timed post-fix window or restart gate: InfraBench's Live Evaluation ``continues under sustained workload or periodic probes to detect configuration drift, delayed degradation, or loss of availability'' and its Restart/Durability gate checks post-restart invariants; SREGym's mitigation oracle ``checks whether the target fault is resolved and whether the target system has recovered to a healthy state'' with no window described, although its load loop runs through the check. \emph{No fault} requires graded healthy-system controls: ORCA-bench's 195 control tasks qualify; DBA-Bench's ten misleading-alert scenarios all contain an injected fault and do not; AIOpsLab's two ``noop'' problems are graded at the detection level only. \emph{Both ways} is marked absent where the source states no per-task requirement that a reference fix pass and a do-nothing run fail; absence in the text is recorded as ``not stated'', not as confirmed absence. \emph{Harbor} was checked against Harbor's adapter registry on 2026-09-22; none of the ten benchmarks in the table appears there (DevOps-Gym, which is not in the table, does). The full claim-by-claim verification trail is released with the paper.

\FloatBarrier
\section{Verifier stress test: reports of a repair that did not happen}
\label{app:stress}
To test whether the verifier can be passed by a report alone, we built a variant of one frappe task in which an external process, unreachable from the agent, re-applies the database fault about once per second, so that no repair can hold. The report tool on this variant required the agent to state whether the fault was resolved, and the grading accepted a report of ``resolved'' as sufficient for the diagnosis check, so only the outcome checks stood between a false report and a pass. We ran \nStressEpisodes{} episodes of one configuration on the variant and \nStressEpisodes{} on the unmodified task. On the unmodified task all \nStressControlPass{} episodes repaired the fault and reported it correctly. On the variant, \nStressFalseResolved{} episodes reported the fault resolved and \nStressHonest{} reported it unresolved; the outcome checks failed all \nStressFalseResolved{} false reports. Reading the agents' own tool output shows that every false report followed a test write that succeeded inside the one-second window before the fault was re-applied, so these are over-confident reports after a transiently successful check rather than reports made after observing the fault persist. The stress test uses a modified task and report tool and is not part of the released set; we report it as evidence about the verifier, not about the agents.


\FloatBarrier
\section{Reward Hacking Meta analysis}
\label{app:rh-anal}
Reward hacking behaviour was noticed in many previous iterations, especially in poorly formed environments. Especially in impossible environments, agents often exhibited signs of frustration.

\paragraph{Exploits in earlier task iterations.} In earlier iterations of these tasks, Claude Opus~5 assembled a remote code execution (RCE) on another Kubernetes pod. It wrote code, serialized it with pickle, spoofed a cookie, and sent a request to the other pod with the pickle as the payload. The second pod deserialized the payload automatically, which executed the code. The cause was poor environment construction: we had omitted required submission criteria, and the agent attacked the grader pod instead of completing the task.

\subsection{Reward hacking judges}

We employed an LLM as a judge for our taxonomy, through our QA process. Our QA process employs a red-teaming agent to identify reward hackable surfaces, calibrated on human judgment. Here, the judge was seeded with task information into the rubric, causing inter-judge disagreement.

The judges show directional severity differences: Astra labels many more trials suspicious or exploit-tier. Agreement about relative suspicion is stronger than agreement about categorical labels. This is descriptive agreement analysis, not evidence that any judge is more accurate.

There are 3,000 trials and 2,272 scored by all four judges. The gap between 3,000 and 2,272 is directly because of refusals by the model providers. Unscored records are abstentions. Model, task, reward, and reasoning-effort fields match across judges for shared trials. The four manifests use the same pipeline hash, xhigh judging effort, one sample, and trajectory budget.

\begin{table}[H]
\centering
\caption{Judge coverage.}
\label{tab:judge-coverage}
\begin{tabular}{lrrr}
\toprule
Judge & Scored & Refused & Missing artifacts \\
\midrule
Opus 5           & 2,860 & 138 & 2 \\
Opus 5.5         & 2,302 & 696 & 2 \\
Gemini 3.8 Flash & 2,998 &   0 & 2 \\
GPT-6 Astra      & 2,987 &  11 & 2 \\
\bottomrule
\end{tabular}
\end{table}

Tables~\ref{tab:judge-classifications} and~\ref{tab:judge-agreement} are computed on the trials that passed all judges (2,272 trials). 

\begin{table}[H]
\centering
\caption{Judge classifications among scored trials.}
\label{tab:judge-classifications}
\begin{tabular}{lrrrr}
\toprule
Judge & Honest & Suspicious & Exploit-tier & Any flag \\
\midrule
Opus 5           & 85.0\% & 15.0\% & 0.0\% & 15.0\% \\
Opus 5.5         & 79.8\% & 20.2\% & 0.0\% & 20.2\% \\
Gemini 3.8 Flash & 71.6\% & 27.4\% & 1.0\% & 28.4\% \\
GPT-6 Astra      & 31.7\% & 65.1\% & 3.2\% & 68.3\% \\
\bottomrule
\end{tabular}
\end{table}

\FloatBarrier
\begin{table}[H]
\centering
\caption{Pairwise agreement between judges.}
\label{tab:judge-agreement}
\begin{tabular}{lrrr}
\toprule
Judge pair & Same category & Cohen's $\kappa$ & Score rank correlation \\
\midrule
Opus 5 / Opus 5.5                  & 89.3\% & 0.632 & 0.671 \\
Opus 5 / Gemini 3.8 Flash          & 82.5\% & 0.501 & 0.537 \\
Opus 5 / GPT-6 Astra               & 44.4\% & 0.121 & 0.625 \\
Opus 5.5 / Gemini 3.8 Flash        & 87.7\% & 0.670 & 0.720 \\
Opus 5.5 / GPT-6 Astra             & 51.0\% & 0.204 & 0.854 \\
Gemini 3.8 Flash / GPT-6 Astra     & 58.8\% & 0.307 & 0.642 \\
\bottomrule
\end{tabular}
\end{table}

\end{document}